\documentclass{article}

\usepackage{PRIMEarxiv}

\usepackage[utf8]{inputenc} 
\usepackage[T1]{fontenc}    
\usepackage{hyperref}       
\usepackage{url}            
\usepackage{booktabs}       
\usepackage{amsfonts}       
\usepackage{nicefrac}       
\usepackage{microtype}      
\usepackage{lipsum}
\usepackage{fancyhdr}      
\usepackage{graphicx}

\usepackage{times}
\usepackage{epsfig}
\usepackage{graphicx}
\usepackage{amsmath}
\usepackage{amssymb}
\usepackage{multirow}
\usepackage{booktabs}

\usepackage{caption} 
\graphicspath{{media/}}

\title{MLGCN: An Ultra Efficient Graph Convolution Neural Model for 3D Point Cloud Analysis
}

\author{
  Mohammad Khodadad$^{\dagger}$\\
  Faculty of Engineering\\
  McMaster University\\
  \And
  Morteza Rezanejad$^{\dagger}$\\
  Department of Psychology\\
  University of Toronto
  \thanks{Dr. Morteza Rezanejad contributed to this article in his personal capacity as an adjunct researcher at the University of Toronto.
    \newline
 \indent ~$^{\dagger}$These authors made equal contributions to this article.\newline
\indent\indent Corresponding author: Morteza Rezanejad (\url{morteza.rezanejad@utoronto.ca}).
}
  \AND
  Ali Shiraee Kasmaee\\
  Faculty of Engineering\\
  McMaster University\\
  \And
  Kaleem Siddiqi\\
  School of Computer Science\\
  McGill University\\
  \AND
  Dirk Walther\\
  Department of Psychology\\
  University of Toronto\\
  \And
  Hamidreza Mahyar\\
  Faculty of Engineering\\
  McMaster University\\
}

\begin{document}
\maketitle

\begin{abstract}
The analysis of 3D point clouds has diverse applications in robotics, vision and graphics. Processing them presents specific challenges since they are naturally sparse, can vary in spatial resolution and are typically unordered. Graph-based networks to abstract features have emerged as a promising alternative to convolutional neural networks for their analysis, but these can be computationally heavy as well as memory inefficient.
To address these limitations we introduce a novel Multi-level Graph Convolution Neural (MLGCN) model, which uses Graph Neural Networks (GNN) blocks to extract features from 3D point clouds at specific locality levels. Our approach employs precomputed graph KNNs, where each KNN graph is shared between GCN blocks inside a GNN block, making it both efficient and effective compared to present models. We demonstrate the efficacy of our approach on point cloud based object classification and part segmentation tasks on benchmark datasets, showing that it produces comparable results to those of  state-of-the-art models while requiring up to a thousand times fewer floating-point operations (FLOPs) and having significantly reduced storage requirements. Thus, our MLGCN model could be particular relevant to point cloud based 3D shape analysis in industrial applications when computing resources are scarce.

\end{abstract}

\section{Introduction}

With advances in 3D acquisition technologies, 3D sensors are becoming more accessible and cost-effective. Sensors including 3D scanners, LiDARs, and RGB-D cameras (e.g., RealSense, Kinect, and Apple depth cameras) provide a wealth of information about the shape, scale, and geometry of objects in the environment. Consequently, there has been an increasing need to develop algorithms and models for point cloud analysis and 3D model classification and segmentation have become active areas of research in machine learning and computer vision. Deep learning techniques have proven to be highly effective for this task due to their ability to learn rich features and representations from raw data. However, most existing 3D deep learning models rely on large and complex architectures, making them computationally expensive and unsuitable for real-time applications, such as augmented reality, robotics, and autonomous driving.

Most sensors on modern 3D perception devices acquire data in the form of point clouds and, traditionally, researchers sample this data on voxel grids for 3D volumetric convolutions. However, the use of low-resolution can result in information loss, e.g., when multiple points fall within the same voxel. To preserve  necessary detail in the input data, a high-resolution representation is preferable, but this can lead to an increase in computational costs and memory requirements.
Whereas data acquired by sensors is often in the form of 3D point clouds, they are unordered and sparse, requiring models 
that are permutation agnostic and multi-scale.
Whereas classical Convolution Neural Network (CNN) models have been effective for image-based computer vision problems, they cannot be directly applied to 3D point cloud analysis.

\begin{figure*}[!ht]
    \centering
    \begin{tabular}{c}
      \includegraphics[width = 0.847\textwidth]{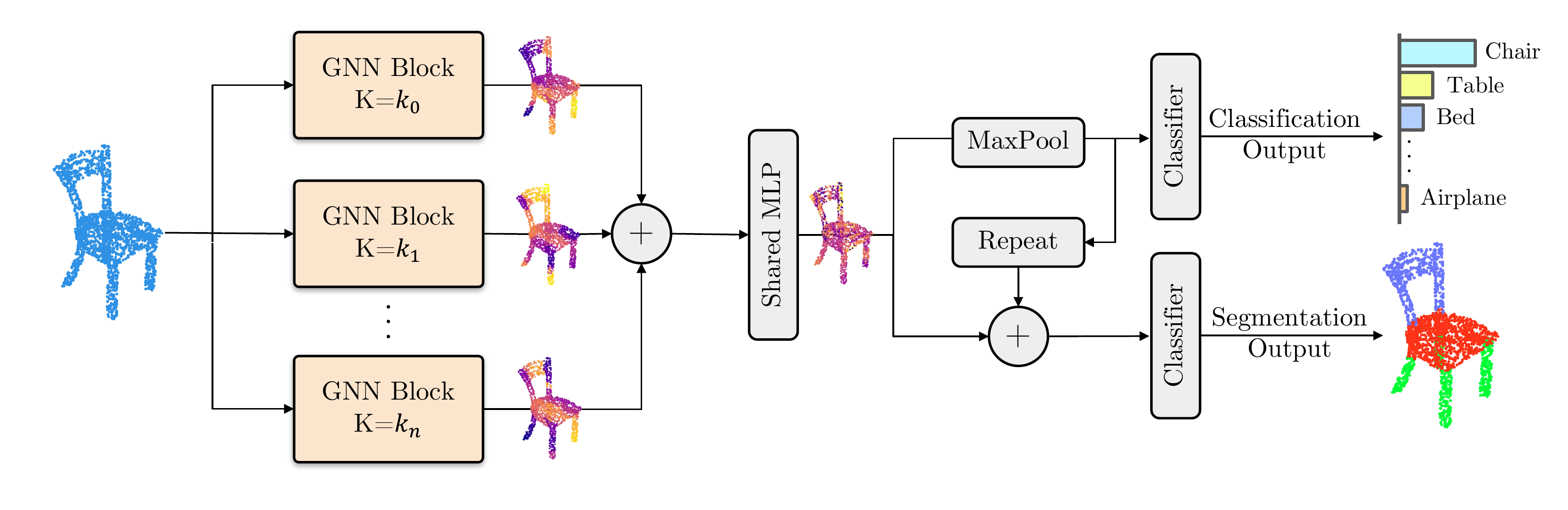} \\
      \includegraphics[width = 0.9\textwidth]{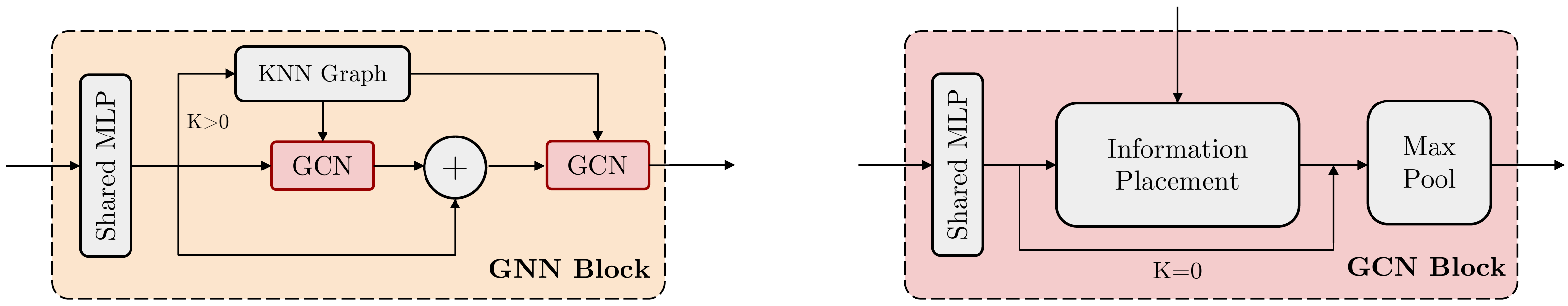} \\
    \end{tabular}
    \caption{Top: The 
    overall architecture of our 3D point cloud processing model, which is designed to be lightweight and efficient for deployment on low-memory, low-CPU devices. Points sampled from an object are fed to GNN blocks for computing features at various spatial locality levels, which are subsequently used for downstream tasks. Bottom: the design of our GNN and GCN blocks, which are the building blocks of our proposed 3D object processing model. One can include as many GCN blocks as needed, where `+' denotes the concatenation operation. }
    \label{fig:design}
\end{figure*}

In recent years, numerous powerful models have been proposed to  analyze point clouds  \cite{qi2017pointnet++,qiu2021geometric,rezanejad2022medial,ma2022rethinking,wang2019dynamic,chen2019gapnet,wan2021dganet,huang2022dual}. Most of these models, however, suffer from a significant drawback: they are typically too complex in terms of parameters and require a large number of mathematical operations, making them unsuitable for industrial use or deployment on lightweight compute devices. Specifically, many of them need to calculate graphs of connectivity on top of point clouds multiple times, resulting in a large number of Floating Point Operations (FLOPs).

Our work addresses the above limitations by introducing a lightweight model that can be trained easily and deployed on low-memory and low-end CPU devices. Instead of relying on complex structures, such as attention mechanisms or deep stacks of feature extraction blocks, which require a large amount of training data and are susceptible to over-fitting, our proposed model (see Figure \ref{fig:design}) consists of multiple shallow graph-based network blocks that capture information from point clouds using different graph KNNs. The use of different KNN graphs combined with shallow GNNs can alleviate the over-smoothing issue caused by deep GCNs \cite{li2018deeper, zhou2020graph}. Furthermore, utilizing precomputed shared graph KNNs, within a GNN block, greatly reduces the number of floating point operations. This architecture offers an efficient solution for processing point clouds without compromising accuracy, making it practical for real-world applications. Our paper 
makes the following contributions:
\begin{enumerate}
    \item We propose a multi-branch graph-based network that effectively captures features at various spatial locality levels of 3D objects, using efficiently designed and lightweight graph-based neural network blocks.
    \item We should that our models are significantly more efficient both in terms of computation and storage, than exsiting approaches for abstracting features from graphs for downstream computer vision tasks.
    \item We conduct a series of ablation studies to analyze the impact of different branches on our model's performance, to gain insight into the role that specific branches play.
\end{enumerate}

\section{Related Work}

Over the past few years, the field of deep learning has seen a surge in research efforts aimed at developing effective methods for analyzing sensor data. Methods that are designed for 2D images cannot be directly applied to 3D point clouds which can be sparse, nonuniform in density and lack local spatial ordering. A promising neural network models for 3D shape analysis in this setting is the PointNet model \cite{qi2017pointnet}. Unlike  previous methods that transform point cloud data to regular 3D voxel grids or collections of images, PointNet processes point cloud data directly, extracting information from individual points and aggregating this it into a feature vector using Global Max Pooling. 

The PointNet model's inability to capture local structures induced by the metric space limits its ability to represent fine-grained patterns and also generalize to complex scenes. To address this issue, PointNet++ \cite{qi2017pointnet++} applies PointNet recursively on nested partitions to extract local features, then combining the learned features across multiple scales. 

In \cite{qiu2021geometric}, the GBNet combines channel affinity modules and CNN networks to improve the representation of point clouds, while learning both local and global features. GBNet utilizes an error-correcting feedback structure to design a back-projection CNN module, 
In \cite{rezanejad2022medial}  medial spectral coordinates are added as additional features to point cloud 3D coordinates. These coordinates contain both local and global features, resulting in improved performance of vanilla models for computer vision tasks. 

The PointMLP model \cite{ma2022rethinking} utilizes MLPs to gather local information from points in a hierarchical manner without using a local feature extractor. Additionally, it employs lightweight affine modules to transform information from points to a normal distribution.

GNNs have the unique ability to handle topologically-structured data without requiring explicit encoding into vectors, by capturing graph-based information \cite{Scarselli2009TheGN}, making them an ideal candidate for the efficient processing of point clouds. The authors of \cite{wang2019dynamic} proposed the DGCNN model where an EdgeConv neural network module incorporates local information around each point, and is then stacked to learn global shape properties using Graph Convolutional Networks.

Zhang and colleagues \cite{zhang2019linked} enhanced DGCNN by introducing LDGCNN, which links hierarchical features from different dynamic graphs to calculate informative edge vectors. 
They removed the transformation network from DGCNN and showed that an MLP can extract transformation-invariant features. They further improved performance by freezing the feature extractor and retraining the classifier.

As attention mechanisms gained momentum in capturing node representation on graph-based data, Chen and colleagues proposed the GAPNet model \cite{chen2019gapnet} which embeds a graph attention mechanism within stacked MLP layers to learn local geometric representations. The GAP-Layer employs an attention-based graph neural network to consider the importance of each neighbor of a point.

The DGANET model \cite{wan2021dganet} uses an improved KNN search algorithm to construct a local dilated graph for each point, modeling long-range geometric correlations with its neighbors. This helps the point neural network to learn more local features of each point, with a larger receptive field during the convolution operation. The authors embed an offset-attention mechanism into a dilated graph attention module and employ graph attention pooling to aggregate the most significant features.  

Huang and colleagues \cite{huang2022dual} propose the Dual-Graph Attention Convolution Network (DGACN), which introduces an improved version of graph attention that leverages information from different hops. They also propose a novel graph self-attention mechanism that extracts more informative features from point clouds.

The Point-transformer \cite{Zhao_2021_ICCV} model utilizes self-attention to capture local information in the vicinity of each point. In addition, the authors introduce a trainable positional encoding that is learned within the end-to-end network.
\cite{wan2021dganet, huang2022dual, wang2019graph, Zhao_2021_ICCV} use graph attention-based mechanisms that are known to be parameter-heavy and can make the process of training and inference computationally expensive.

\section{The Proposed Method: MLGCN}
MLGCN is a multi-level graph neural network model that can capture information from 3D point clouds at different locality levels efficiently. The model consists of multiple GNN blocks, each taking a set of 3D point clouds as input and learning a representation of the 3D dataset.  
The model then concatenates and uses these features for downstream tasks. We have designated two downstream branches: one for a classification task (i.e., correctly labeling the 3D model), and one for a segmentation task (i.e., decomposing the model into a set of semantically meaningful parts). In this section, we describe the key components of the MLGCN model, a schematic of which is shown in Figure \ref{fig:design}. We assume the following point cloud as input to the system:
\begin{equation}
\mathcal{X} =  \left \{ \mathbf{p}_i = (x_i,y_i,z_i) \in \mathbb{R}^3\text{ for }i =  1, 2, \cdots, N \right \}
\end{equation}

\subsection{KNN Graphs}
Given 3D point cloud data, the model forms a set of KNN graphs, where nodes represent 3D points, and each node is connected to its $k$ closest nodes using edges. The parameter $k$ defines the locality level around each point where local neighborhood information will be collected. The unique utilization structure of our KNN graph is that the graph is computed once for an input $\mathcal{X}$ and then reused for various other blocks' outputs of $Y$. This approach saves computation time and resources, making our model very efficient. The edge connectivity from the KNN graph is used to decide on passing information (messages) over an edge, allowing the model to capture the global features of the input data. Overall, the KNN graph used in our MLGCN model provides a way to explore the local structure of 3D point clouds as well as capture global features efficiently and effectively.
To formulate the KNN graph construction, we define the graph $\mathcal{G}_{k}$ as:
\begin{equation}
    \mathcal{G}_{k} = (\mathcal{X},E_{k})
\end{equation}
where $\mathcal{X}$ represent the nodes in our graph and $E_{k} \subseteq \mathcal{X} \times \mathcal{X}$ represents the edges. Each node $\mathbf{p}_i$ is connected to another node $\mathbf{p}_j$ if $\mathbf{p}_j$ locates within the $k$ closest neighbors of $\mathbf{p}_i$. As the graph is directed, the graph contains self-loops (see Figure \ref{fig:design} bottom left).

\subsection{GNN Block}
Each Graph Neural Network (GNN) block takes a 3D point cloud as input and extracts features from it. These features are then concatenated and used for both classification and segmentation tasks.
To extract these features, the GNN block applies a series of operations on the input data. First, a multi-layer perceptron (MLP) is applied to transform the input, which is then processed by a series of Graph Convolution Network (GCN) blocks and one single KNN graph. If the parameter $k$ is set to $0$, the model skips the KNN graph computation and only extracts global information from the point cloud.

Each GCN block processes the input data and then its output is concatenated with its input and passed to the next GCN block, along with the output of the KNN graph. The KNN graph output is shared between GCN blocks in a GNN block. The next GCN block operates similarly to the previous one, processing the concatenated features to extract additional information. This process can be repeated multiple times except for the last GCN block where the input and output vectors are no longer concatenated. In Figure \ref{fig:design} bottom left, we illustrate the GNN block architecture. Here

\begin{equation}
    \Gamma(\mathcal{X}) = f\left( \text{concat} \{ GB_{k_i}(\mathcal{X}) | i = 1, \cdots, m\} \right),
\end{equation}
where $f$ is the shared MLP applied to the concatenated outputs of the GNN blocks $GB(\mathcal{X})$.
\begin{table*}[!ht]
    \centering
    \begin{tabular}{c@{\hskip 3pt}c@{\hskip 3pt}c@{\hskip 3pt}c@{\hskip 3pt}c@{\hskip 3pt}c@{\hskip 3pt}c@{\hskip 3pt}c@{\hskip 3pt}c}
        \toprule
        { \multirow{2}{*}{Method}} & Input    & { Model Size} & { FLOPS} & { Number of parameters} & { \multirow{2}{*}{Accuracy}} & { GPU Memory} \\
        & Shape & { Mega Bytes} & {(100 Mega)} & {  100 Thousands} & & { Mega Bytes} \\
        \hline
        { Pointnet (vanilla)
        \cite{qi2017pointnet}
        } & 1024 & - & 1.5 & 8 & 87.1 & -
        \\
        { Pointnet
        \cite{qi2017pointnet}
        } & 1024 & 38 & 4.5 & 35 & 89.2 &  50
        \\
        { Pointnet++ 
        \cite{qi2017pointnet++}
        } & 1024 & 17 & 8.9 & 14 & 90.7 & 100
        \\ 
        { GBNet 
        \cite{qiu2021geometric}
        } & 1024 & 34 & 98 & 87 & 93.8 & 220
        \\
        { PointMLP 
        \cite{ma2022rethinking}
        } & 1024 & 100 & 157 & 132 & \textbf{94.5} & 90
        \\\hline
        { DGCNN 
        \cite{wang2019dynamic} 
        } & 1024 & 21 & 1300 & 18 & 92.9 & 110 
        \\
        { LDGCNN
        \cite{zhang2019linked}
        } & 1024 & 13 & 920 & 10 & 92.9  & -
        \\
        { DGANET 
        \cite{wan2021dganet}
        } & 1024 & 6 & - & 15 & 92.3 & -
        \\
        { GAPNet 
        \cite{chen2019gapnet}
        } & 1024 & 21 & 580 & 19 & 92.4 & 31
        \\
        { DGACN 
        \cite{huang2022dual} 
        } & 1024 & - & 1600 & 240 & 94.1 & -
        \\
        { Point-Transformer 
        \cite{Zhao_2021_ICCV} 
        } & 1024 & 82 & - & 140 & 93.7 & 155   
        
        \\\hline
        { Light MLGCN} & 1024 & \textbf{1.5} & \textbf{1.3} & \textbf{1.2} & 90.7 & 45
        \\
        { Lighter MLGCN} & \textbf{512} & \textbf{0.4} & \textbf{0.2} & \textbf{0.3} & 88.6 & \textbf{10} \\
        \bottomrule
    \end{tabular}
    \caption{We carry out a comparison of various models using different metrics, with processing on the ModelNet-40 dataset as the basis for evaluation.}
    \label{tab:cscore}
\end{table*}

\subsection{GCN block}
The GCN block in our MLGCN model applies a series of operations on the input data using the KNN graph information that was computed previously. The input data is first processed by a shared multi-layer perceptron. The GCN block then uses the KNN graph information to propagate the input feature information for each node and the nodes it is connected to. This operation allows the model to capture local features of the input data using the precomputed KNN graph. The output of the GCN block is then max pooled. This max pooling operation summarizes the information learned from the input data and allows the model to capture the most important features of the input with respect to the defined locality level $k$.

Our information placement module uses graph connectivity as follows. We assume our message passing function $h(\mathbf{p}_i,\mathbf{p}_j,Y)$ accepts two nodes $\mathbf{p}_i,\mathbf{p}_j$ and then passes the information ($y_j \in Y$) on node $\mathbf{p}_j$ to node $\mathbf{p}_i$ conditioned on the graph neighborhood information, i.e., if $(\mathbf{p}_i,\mathbf{p}_j) \in E_{k}$. Here $E_{k}$ is shared among all GCN blocks that belong to the same GNN block. In Figure \ref{fig:design} bottom right, we show the GCN block architecture.

\subsection{Information Processing in the GCN Block}
As mentioned previously, a GNN block input is a 3D point cloud $\mathcal{X}$ where a graph $\mathcal{G}_{k} = (\mathcal{X},E_{k})$ is made. We now explain how the inputs and outputs of each GCN block are obtained.  Let  $y_{i}^{t}$ represent the information from the $i^{th}$ node of our graph after the $t^{th}$ GCN block operation is applied on the input. We can formulate $y_{i}^{t}$ as
\begin{equation}
y_{i}^{t} = A\left( \left \{ h(\mathbf{p}_{i},\mathbf{p}_{j},f_{t}(Y^{*(t-1)})) | (\mathbf{p}_{i},\mathbf{p}_{j}) \in E_{k} \right \} \right)
\end{equation}
where $A$ is the aggregation function and $f_t$ is the $t^{th}$ shared MLP. The aggregation function used in our pipeline is max pooling (athough other aggregation functions could be used as well) and it is applied along the neighborhood axes. For all GCN blocks except for the last one, the information is $y_{i}^{t}$ concatenated with the input to the same GCN block:
\begin{equation}
    y_{i}^{*t} = \text{concat}\left(y_{i}^{t},y_{i}^{*(t-1)}\right).
\end{equation}
For $t = 1$, $y_{i}^{t} = y_{i}^{*t} = f_0(\mathcal{X})$. 
Now, with the GNN block represented by $GB(\mathcal{X})$, 
$
GB(\mathcal{X}) = Y^{l}
$
where $l$ is the index of the last GCN block.

\subsection{Overall Architecture}
Each variation of MLGCN uses a set of GNN blocks with different values of $k$. Let the first block be a block with $k = 0$, with the purpose of extracting global information for each node. The other blocks can be set to extract information with different locality levels. Now assume we have a set of $m$ different GNN blocks in our model with $K = \{ k_1 , k_2, \cdots, k_m\}$. As mentioned previously, outputs of all GNN blocks are concatenated and then passed through a shared MLP. From there, the extracted features are pooled and then used in a downstream task, e.g., classification or segmentation.

\subsubsection{Classification Branch}
We designated a classification branch to classify 3D input models according to different labels. For the classification task, we simply apply a max pooling along the node's axes and pass the outcome to a classifier as follows:
\begin{equation}
\mathcal{L}_{\text{classification}} = \mathcal{C}\left(A\left(\Gamma(\mathcal{X})\right)\right)    
\end{equation}
where $\mathcal{L}_{\text{classification}}$ is the set of classification labels, $\mathcal{C}$ is a classifier and $A$ is the max pooling function here.

\begin{table*}[!ht]
\renewcommand{\arraystretch}{1.3}
\begin{footnotesize}
    \centering
    \begin{tabular}{c@{\hskip 2pt}|@{\hskip 2pt}c@{\hskip 2pt}|@{\hskip 2pt}c@{\hskip 2pt}|@{\hskip 2pt}c@{\hskip 2pt}c@{\hskip 5pt}c@{\hskip 2pt}c@{\hskip 5pt}c@{\hskip 2pt}c@{\hskip 5pt}c@{\hskip 2pt}c@{\hskip 5pt}c@{\hskip 2pt}c@{\hskip 5pt}c@{\hskip 2pt}c@{\hskip 5pt}c@{\hskip 2pt}c@{\hskip 5pt}c@{\hskip 2pt}c}
        \toprule
        \multirow{2}{*}{Method} &  Class  &   Inst. & \multirow{2}{*}{aero} & \multirow{2}{*}{bag} & \multirow{2}{*}{cap} & \multirow{2}{*}{car} & \multirow{2}{*}{chair} & ear- & \multirow{2}{*}{guitar} & \multirow{2}{*}{knife} & \multirow{2}{*}{lamp} & \multirow{2}{*}{laptop} & motor- & \multirow{2}{*}{mug} & \multirow{2}{*}{pistol} & \multirow{2}{*}{rocket} & skate- & \multirow{2}{*}{table} 
        \\& mIoU & mIoU & & & & & & phone & & & & & bike &  & & & board &  \\
        \hline
        {Pointnet 
        } & 80.4 & 83.7 & 83.4 & 78.7 & 82.5 & 74.9 & 89.6 & 73.0 & 91.5 & 85.9 & 80.8 & 95.3 & 65.2 & 93.0 & 81.2 & 57.9 & 72.8 & 80.6
        \\
        {Pointnet++ 
        } & 81.9  & 85.1 & 82.4 & 79.0 & 87.7 & 77.3 & 90.8 & 71.8 & 91.0 & 85.9 & 83.7 & 95.3 & 71.6 & 94.1 & 81.3 & 58.7 & 76.4 & 82.6
        \\

        {GBNet 
        } &  82.6 & 85.9 & 84.5 & 82.2 & 86.8 & 78.9 & \textbf{91.1} & 74.5 & 91.4 & 89.0 & 84.5 & 95.5 & 69.6 &  94.2 & 83.4 & 57.8& 75.5 & 83.5
        \\
        
        {PointMLP 
        } &  \textbf{84.6} & \textbf{86.1} & 83.5 & 83.4 & 87.5 & \textbf{80.5} & 90.3 & 78.2 & 92.2 & 88.1 & 82.6 & 96.2 & \textbf{77.5} & \textbf{95.8} & \textbf{85.4} & \textbf{64.6} & 83.3 & 84.3
        \\\hline

        {DGCNN 
        } & 82.3 & 85.2 & 84.0 & 83.4 & 86.7 & 77.8 & 90.6 & 74.7 & 91.2 & 87.5 & 82.8 & 95.7 & 66.3 & 94.9 & 81.1 & 63.5 & 74.5 & 82.6
        \\
        {LDGCNN 
        } & 82.2 & 84.8 & 84.0 & 83.0 & 84.9 & 78.4 & 90.6  & 74.4 & 91.0 & 88.1 & 83.4 & 95.8 & 67.4 & 94.9 & 82.3 & 59.2 & 76.0 & 81.9
        \\
        {DGANET 
        } & 82.6 & 85 & 84.6 & \textbf{85.7} & 87.8 & 78.5 & 91.0 & 77.3 & 91.2 & 87.9 & 82.4 & 95.8 & 67.8 & 94.2 & 81.1 & 59.7 & 75.7 & 82.0
        \\
        {GAPNet 
        } & 82 & 84.7 & 84.2 & 84.1 & \textbf{88.8} & 78.1 & 90.7 & 70.1 & 91.0 & 87.3 & 83.1 & 96.2 & 65.9 & 95.0 & 81.7 & 60.7 & 74.9 & 80.8
        \\\hline
        {MLGCN} & 83.2 & 84.6 & \textbf{87.4}	& 78.2	& 85.6 & 	75.6 &	75.9 &	\textbf{81.1} &	\textbf{93.1} &	\textbf{93.2} &	\textbf{89} &	\textbf{96.4} &	67.5 &	93.7 &	81.8 &	60.6 &	\textbf{85.2} &	\textbf{87.6}
        \\
        \bottomrule
    \end{tabular}
    \caption{A comparison of the results achieved by different models for part segmentation on the ShapeNetPart dataset. The results demonstrate that our proposed model performs comparably to the best-performing models in the literature for part segmentation, and in some cases, even outperforms them. We obtain the best score for 8 out of 16 object classes. }
    \label{tab:sn_scores}
\end{footnotesize}
\end{table*}

\subsubsection{Segmentation Branch}
The second designated branch in our overall architecture is dedicated to the part segmentation of the 3D models. For the segmentation task, the model concatenates the information of each node with the repeated pooled information obtained for all the nodes from the GNN blocks that is used in the classification branch:
\begin{equation}
\mathcal{L}_{\text{segmentation}} = \mathcal{C}\left(\text{concat} \left( \text{repeat}\left(\Gamma(\mathcal{X})\right), \Gamma(\mathcal{X}) \right)\right)
\end{equation}
where $\mathcal{L}_{\text{segmentation}}$ is the set of segmentation labels , $\mathcal{C}$ is a classifier and $A$ is the max pooling function.

\subsection{Light-MLGCN \& Lighter-MLGCN}

Here, we introduce two sample architectures with a MLGCN backbone, Light-MLGCN, and Lighter-MLGCN. These are example models to demonstrate the efficiency of MLGCN-based models. To show this we compare their performance to that of state-of-the-art models that are commonly used for 3D classification and segmentation problems. 

Both Light-MLGCN and Lighter-MLGCN utilize multiple GNN-blocks with varying $k$ sizes. This allows them to capture information related to different locality levels without requiring additional trainable parameters to capture the distance from the neighborhood center. Additionally, the $l$ value for each GNN block is set to 2, resulting in a shallow network that is less susceptible to over-fitting. Moreover, Light-MLGCN computes graphs based on only three features, as the range of $f_{0}$ is 3, which makes its graph calculation process much faster than that of other existing papers. These models share the graph for each GNN-block, which results in fewer mathematical operations. Light-MLGCN was trained using hyperparameters of $K={63,15,0}$, and for each GNN block, $y^0 \in \mathbb{R}^{1024\times3}$, $y^1 \in \mathbb{R}^{1024\times32}$, $y^2 \in \mathbb{R}^{1024\times128}$, and $\Gamma(\mathcal{X}) \in \mathbb{R}^{1024\times256}$. Conversely, Lighter-MLGCN was trained using hyperparameters of $K={31,7,0}$, and for each GNN block, $y^0 \in \mathbb{R}^{512\times3}$, $y^1 \in \mathbb{R}^{512\times16}$, $y^2 \in \mathbb{R}^{512\times64}$, and $\Gamma(\mathcal{X}) \in \mathbb{R}^{512\times128}$.

\section{Experiments}
\label{sec:experiments}
We now evaluate the performance of our MLGCN models with respect to different metrics. We demonstrate that our models achieve comparable accuracy to existing models in both classification and segmentation tasks while being considerably smaller and faster.

\subsection{Implementation Details}
We trained our models on a machine with a single P100 GPU with 12GB memory. For the optimization step, we employed the Adam optimizer, setting the batch size to 128. The initial learning rate was 0.001, which was reduced by a factor of 0.997 ($e^{-0.003}$) after the $20^{th}$ epoch.

\subsection{Classification}

Our primary experiment involves comparing the accuracy and speed of our models on ModelNet-40 \cite{wu20153d}, a dataset consisting of 9,843 training and 2,468 testing meshed CAD models from 40 distinct categories. In Table \ref{tab:cscore}, we compare our model to several recent and popular models in terms of accuracy, floating-point operations, number of trainable parameters, model storage size, and GPU memory.

\begin{figure*}[!ht]
    \centering
    \begin{tabular}{c@{\hskip -15pt}c@{\hskip -8pt}c@{\hskip -3pt}c@{\hskip -9pt}c@{\hskip -5pt}c@{\hskip -5pt}c@{\hskip -1pt}c}
      \includegraphics[height = 0.135\textwidth]{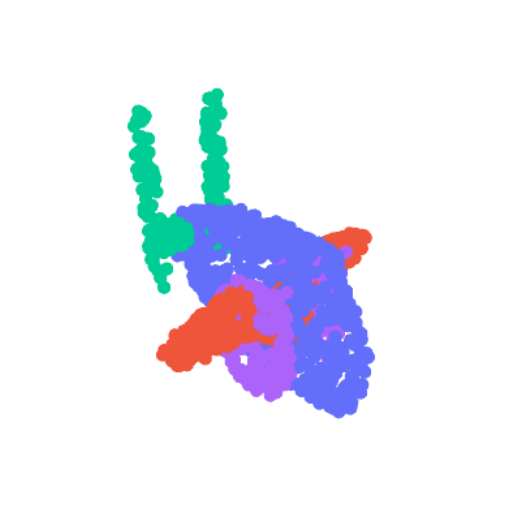} & 
      \includegraphics[height = 0.135\textwidth]{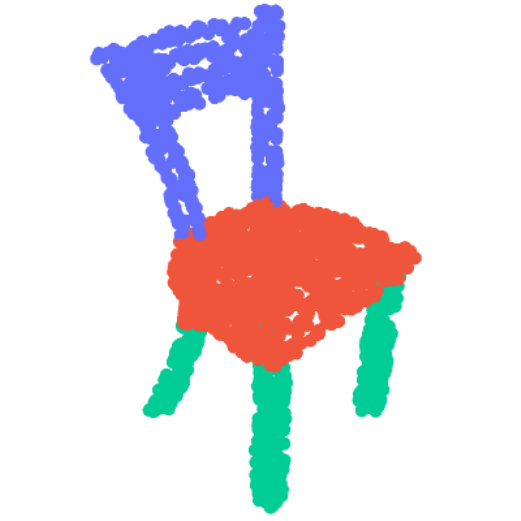} &
      \includegraphics[height = 0.135\textwidth]{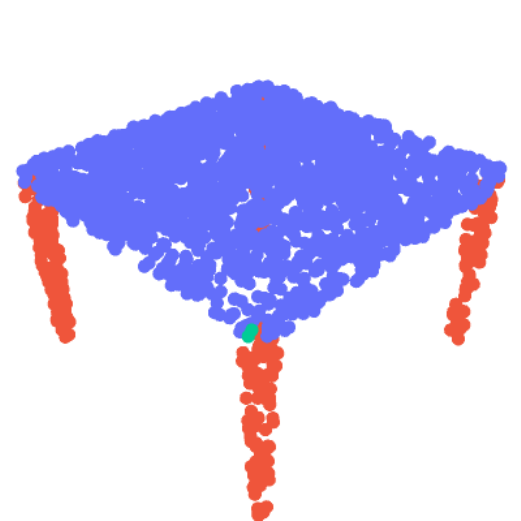} &
      \includegraphics[height = 0.135\textwidth]{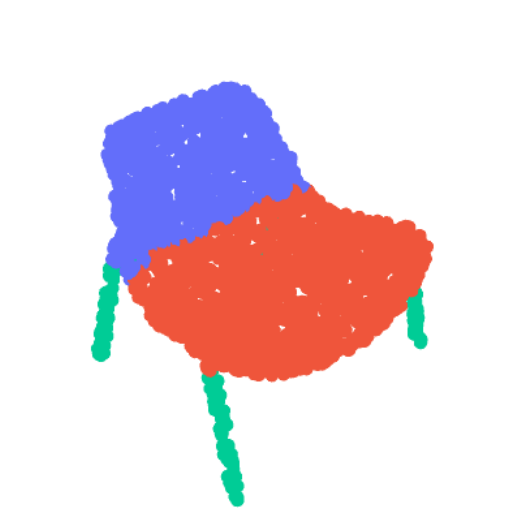} &
      \includegraphics[height = 0.135\textwidth]{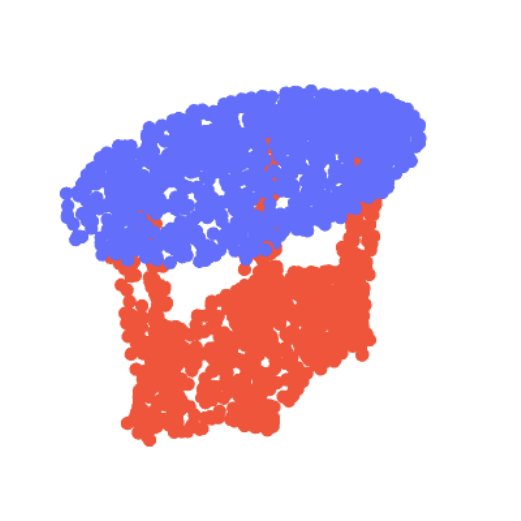} &
      \includegraphics[height = 0.135\textwidth]{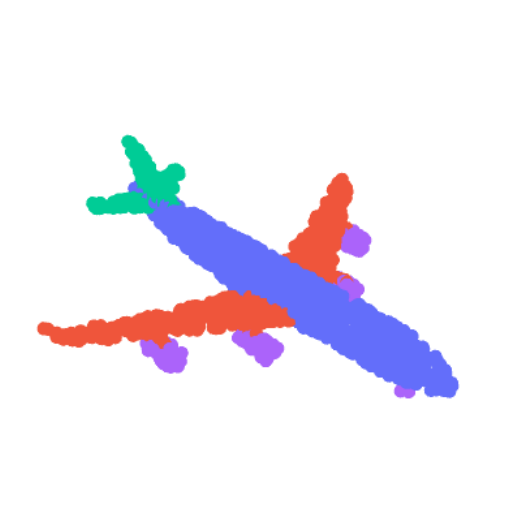} &
      \includegraphics[height = 0.135\textwidth]{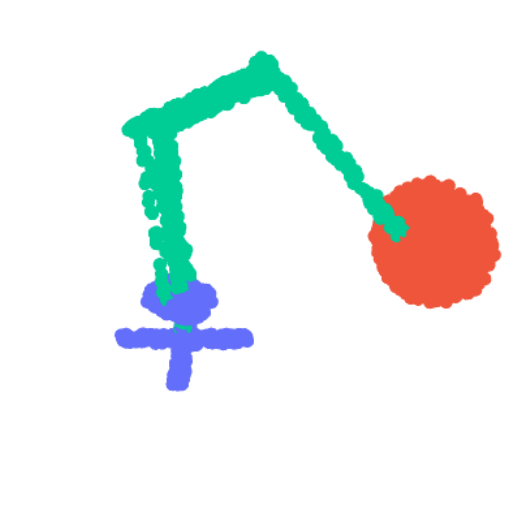} &
      \includegraphics[height = 0.135\textwidth]{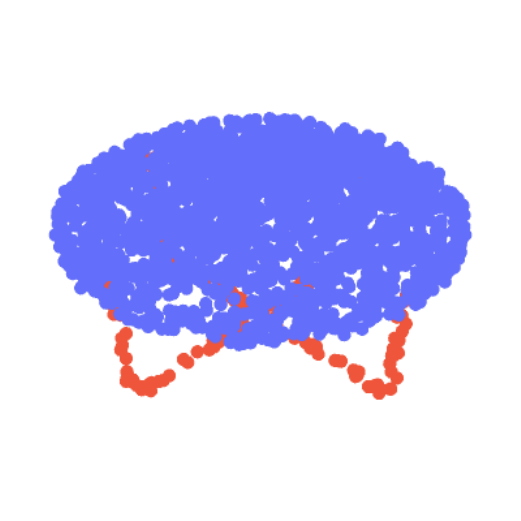}       
      \\
      \includegraphics[height = 0.135\textwidth]{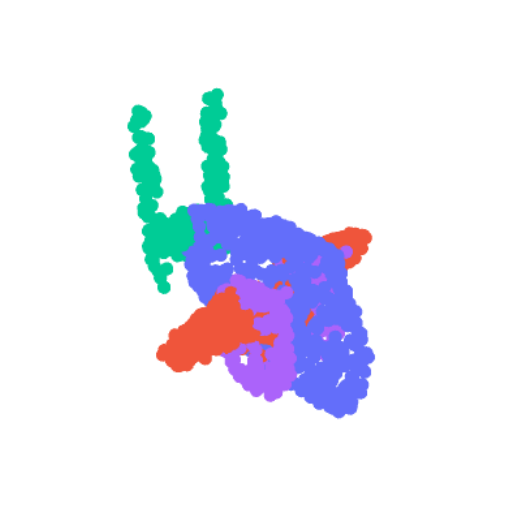} & 
      \includegraphics[height = 0.135\textwidth]{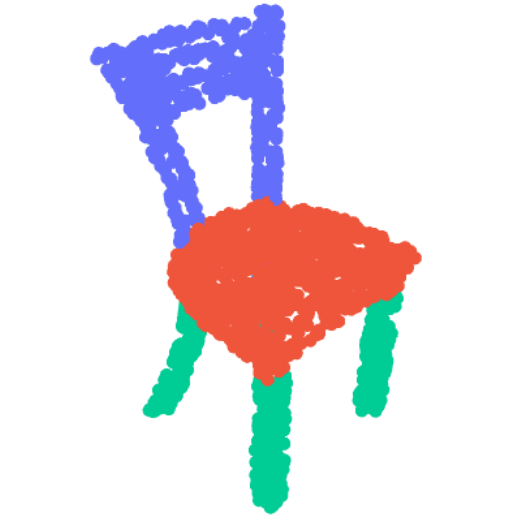} &
      \includegraphics[height = 0.135\textwidth]{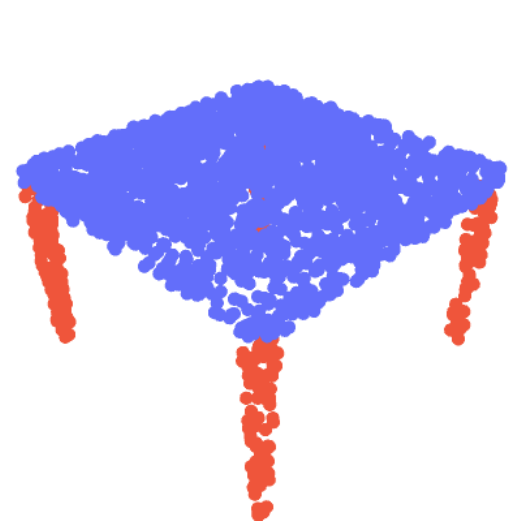} &
      \includegraphics[height = 0.135\textwidth]{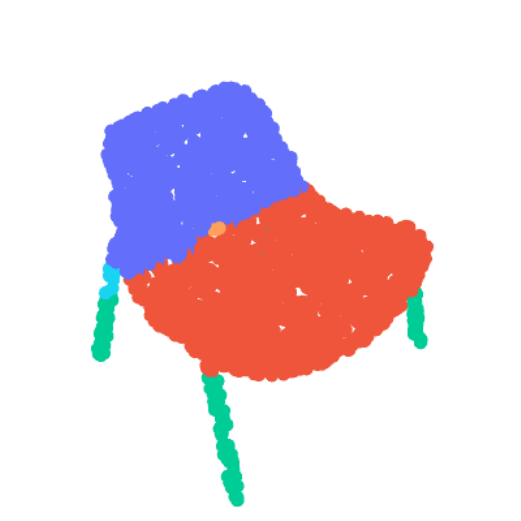} &
      \includegraphics[height = 0.135\textwidth]{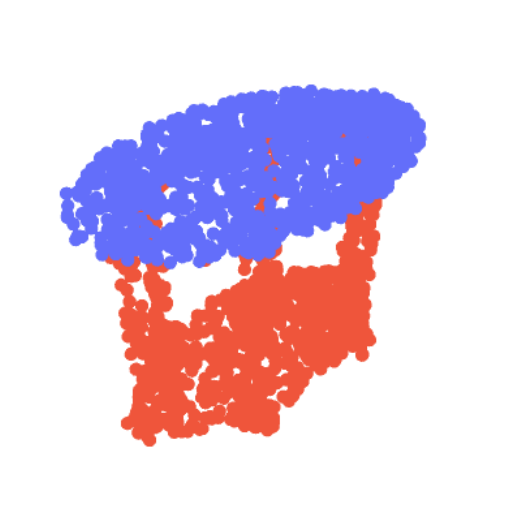} &
      \includegraphics[height = 0.135\textwidth]{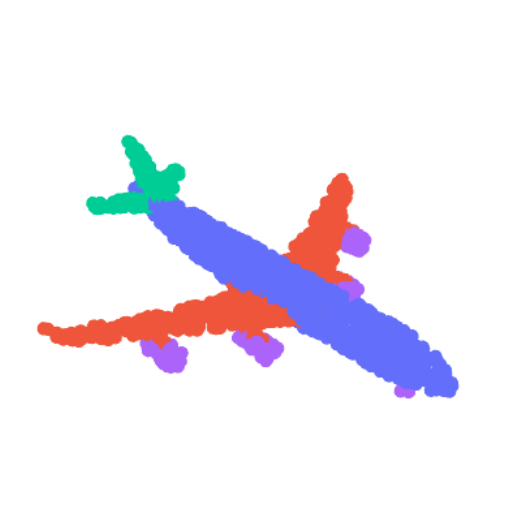} &
      \includegraphics[height = 0.135\textwidth]{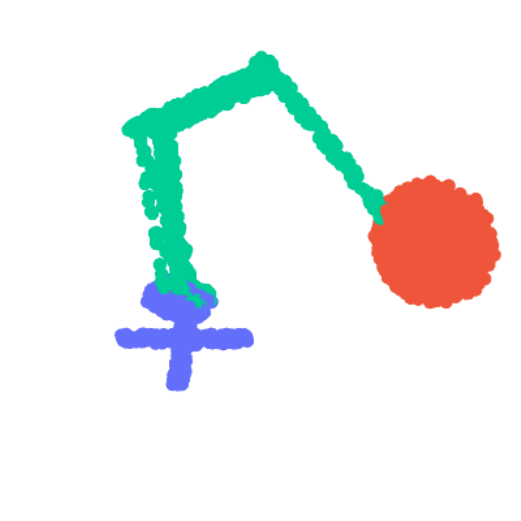} &
      \includegraphics[height = 0.135\textwidth]{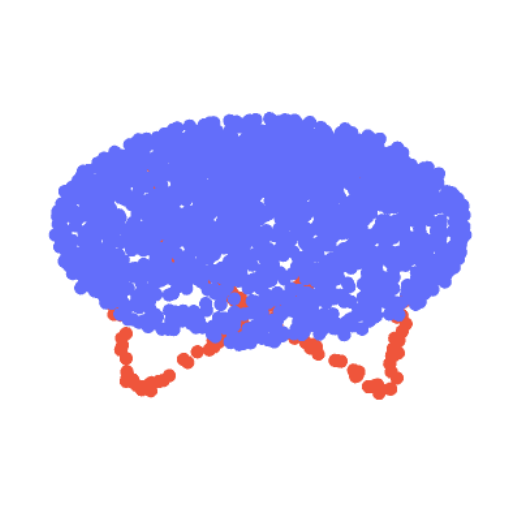}       
    \end{tabular}
    \caption{The top row shows the ground truth segmentation, while the bottom row displays the predicted class output label using our MLGCN model.}
    \label{fig:segres}
\end{figure*}

As shown in Table \ref{tab:cscore}, when comparing Light-MLGCN with the best model in terms of accuracy, we see that it is more than 100 times more efficient in terms of FLOPS, and is also more than 100 times smaller in terms of the number of parameters and more than 60 times smaller in terms of model size. Whereas it has only 3.8 percent lower classification accuracy on the ModelNet-40 dataset than the best model \cite{wu20153d}, Light-MLGCN is considerably faster and more compact. 

Among graph-based models, DGACN achieves the highest accuracy but requires 1230 times more FLOPS than our model while only achieving 3.4 percent higher accuracy. Additionally, Lighter-MLGCN achieves comparable accuracy to Light-MLGCN with only a 2.1 percent difference, while being significantly smaller and faster, and processing only 512 points sampled from point clouds. A detailed presentation of our results is in Table \ref{tab:cscore}.

\subsection{Segmentation}

In addition to the 3D classification problem, we also evaluated the performance of our models on the part segmentation task using the ShapeNetPart dataset \cite{yi2016scalable}. This dataset contains 16,881 3D shapes from 16 different classes, where each class has 2 to 6 parts, resulting in a total of 50 different parts. Our objective is to demonstrate that our lightweight model can achieve comparative results (or even better results) while remaining significantly smaller in size than other existing models. To ensure a fair comparison with previous work, we trained and tested our model on samples comprising 2048 points each, using the same settings as those in other papers. The results are presented in Table \ref{tab:sn_scores}, which shows that our model achieves comparable performance with other state-of-the-art models, despite being much smaller in size. 

Moreover, to provide a visual representation of our model's output, we compared it's output labels to the ground truth in  Figure \ref{fig:segres}. The results show that our model is able to accurately segment the parts of the 3D objects, further demonstrating its efficacy for this task.

\section{Ablation Studies}
We now examine details of our models and demonstrate that they are much more efficient than the other existing models.

\subsection{FLOPS Required for Each Operation}
\label{sec:flops}

In many graph-based models, graph calculation is one of the most computationally intensive operations. To calculate the K-NN graph, the K-nearest neighbor algorithm is used to find the nearest neighbors of each point. This results in a computational complexity of $O(n^2 \times k)$, where $n$ is the number of points and $k$ is the length of the feature vector for each point. This complexity can have a significant impact on the number of floating-point operations required for a graph-based model.

Table \ref{tab:ops} demonstrates that graph calculation can be highly resource-intensive when dealing with a large number of points and features. For instance, the FLOPS required to calculate graphs using KNN can increase dramatically as the number of points and features increase. In contrast, Light-MLGCN employs shared graphs on small feature vectors for multiple GCNs, resulting in reduced computational overhead. As a result, Light-MLGCN is able to achieve comparable performance to other state-of-the-art models while being much faster and smaller in size.

Most current graph-based models used for this specific problem require multiple instances of graph extraction on point clouds with 32 to 128 features. This can result in a large number of floating-point operations, which can lead to reduced performance and longer training times. As shown in Table \ref{tab:cscore}, graph-based models generally require significantly more floating-point operations than non-graph-based models. 

\begin{table}[!t]
\renewcommand{\arraystretch}{1.1}
    \centering
\begin{tabular}{@{\hskip 5pt}c@{\hskip 12pt}c@{\hskip 12pt}c@{\hskip 5pt}}
            \toprule             
             Dimension  & \multirow{2}{*}{Operation Type} &  FLOPS \\
             Configuration & & Mega
            \\\hline\hline
            (1024,3)-(1024,32) & Point-wise Dense & 0.13
            \\
            (1024,32)-(1024,64) & Point-wise Dense & 2
            \\
            (1024,64)-(1024,128) & Point-wise Dense & 8
            \\
            (1024,128)-(1024,256) &Point-wise Dense & 33 
            \\
            (1024,512)-(1024,1024) &Point-wise Dense & 537 
            \\\hline 
            (2048,128)-(2048,256) &Point-wise Dense & 67 
            \\
            (2048,512)-(2048,1024) & Point-wise Dense & 1074 
            \\\hline
            (1024,3) & Graph Calculation & 4
            \\
            (1024,32) & Graph Calculation & 50
            \\
            (1024,64) & Graph Calculation & 100 
            \\
            (1024,128) & Graph Calculation & 201 
            \\
            (1024,512) & Graph Calculation & 805 
            \\\hline 
            (2048,128) & Graph Calculation & 805 
            \\
            (2048,512) & Graph Calculation & 3221 
            \\\bottomrule             
\end{tabular}
\caption{We provide a comparison of the number of floating-point operations (FLOPS) required for different operation types in a model.}
\label{tab:ops}
\end{table}

\subsection{Performance of MLGCN model with Various Input Shapes}
While Light-MLGCN was primarily designed to operate on 1024 points and Lighter-MLGCN on 512 points, both models can be tested on other sampled point cloud sizes. This section aims to demonstrate the effectiveness of our models with different point cloud shapes. We show that our models can perform well even on sparser point clouds. To get a better sense of this, we tested both of our models with input sizes of 128, 256, 512, and 1024 and present the number of FLOPS and their corresponding accuracies in Table \ref{tab:ll}.  

As shown in this table, the simplicity and shallow structure of both Light-MLGCN and Lighter-MLGCN enable them to be trained on smaller point cloud samples without over-fitting, resulting in high accuracy even when using much fewer 3D point cloud sample points. This demonstrates the flexibility of our models and their ability to perform well under varying input conditions.

\begin{table}[!h]
\renewcommand{\arraystretch}{1.1}
    \centering
    \begin{tabular}{cccc}
    \toprule
         Model & Input Shape & FLOPS (Giga) &   Accuracy \\\hline
            \multirow{4}{*}{\rotatebox[origin=c]{90}{Light}}& 1024 & 0.13 & 90.7
            \\
            & 512 & 0.06  & 89.5
            \\
            & 256 &  0.03 &  88.4
            \\
            & 128  &  0.014 &  86.4
            \\\hline\hline
            \multirow{4}{*}{\rotatebox[origin=c]{90}{Lighter}} & 1024 & 0.04 & 89.8 \\
            & 512 & 0.017 & 88.6
            \\
            & 256 & 0.008 &  86.9
            \\
            & 128 & 0.004 & 83.7\\
            \bottomrule
    \end{tabular}
    \caption{The performance of the MLGCN model can vary with different input shapes. In order to evaluate the robustness of the model under different input conditions, we conducted experiments with various input shapes and analyzed the results.}
    \label{tab:ll}
\end{table}

\subsection{MLGCN as an Encoder}
\begin{figure}[!h]
    \centering
      \includegraphics[width = 0.450\textwidth]{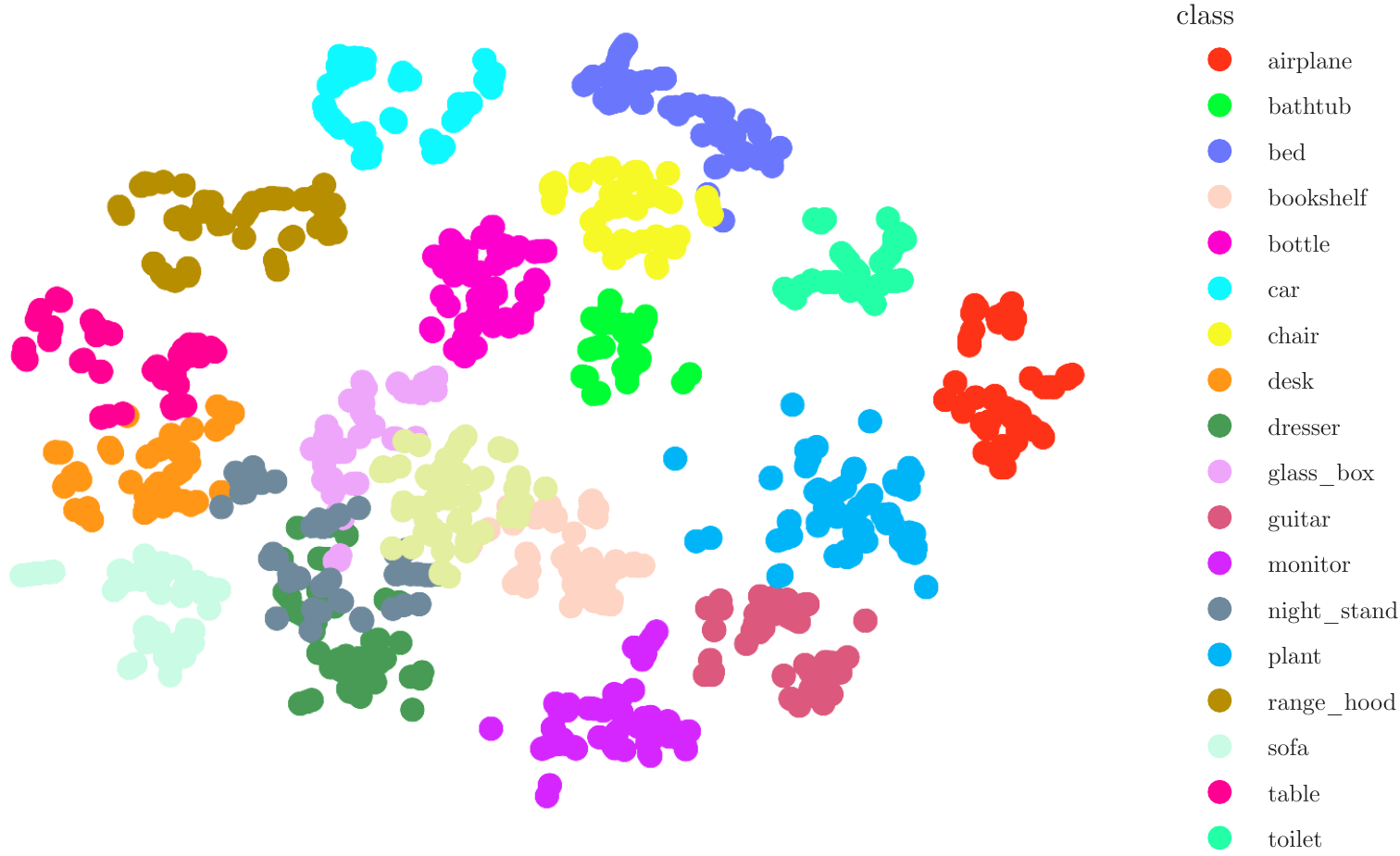}
    \label{fig0}
    \caption{A 2D TSNE plot to visualize the projected features obtained by our proposed model (Light-MLGCN) for 20 different object classes.}
    \label{fig:f_v}
\end{figure}

Our proposed MLGCN model can also serve as an encoder model for encoding 3D point clouds and extracting meaningful features. To evaluate this hypothesis, we extracted the information of the classification $\text{MaxPool}\left(\Gamma(\mathcal{X})\right)$ branch (before the classifier) and projected it into a lower-dimensional space to examine how these features separate between different classes of 3D models. Figure \ref{fig:f_v} presents a (2D TSNE) visualization of the projection of feature vectors generated by our model when tested on the Modelnet-40 dataset onto a 2-dimensional space. The figure clearly demonstrates that our model can effectively cluster each class of 3D objects into a separate cluster, indicating the ability of the model to extract and encode meaningful features from 3D point clouds. It should be noted that Z-score outlier detection was applied to the data. The figure suggests that our proposed model can serve as a robust encoder model for extracting features from 3D point clouds.

\subsection{Role of Different Sets of $K$}
\label{sec:role_k}

In this section, we examine how different sets of $K$ in the GNN blocks of our proposed model impact the accuracy on the Modelnet-40 dataset. We first demonstrate that our selected $K$ values of $[0,15,63]$ perform well in the GNN blocks, as indicated in Table \ref{tab:blocks2}. We observe that our model achieves high accuracy using these $K$ values.

Furthermore, we explore the possibility of combining different $K$ values to improve the accuracy of our model. We do this by using multiple GNN blocks with different $K$ values and concatenating their output features. We discover that incorporating $k=0$, which captures global features, results in a significant improvement in accuracy. 

Finally, it is worth noting that by combining different sets of $K$ values, we can capture multi-scale information with different receptive field sizes, enabling the model to learn both local and global features effectively.

\begin{table}[!ht]
\renewcommand{\arraystretch}{1.1}
    \centering
\begin{tabular}{ccc}
            \toprule
            { Block} & { FLOPS (Giga)} & { Accuracy} \\\hline
            $[0,15,63]$ & 0.13 & 90.7\\ 
            $[0,19,63]$ & 0.13 & 90.4\\ 
            $[0,9,44]$ & 0.13 & 90.3\\     
            $[0,19,44]$ & 0.13 & 90.1\\\hline
            $[15,63]$ & 0.09 & 90.1\\
            $[0, 44]$ & 0.08 &  89.9\\
            $[19, 63]$ & 0.09 &  89.9\\
            $[9, 44]$ & 0.09 &  89.8\\\hline
            $[15]$ & 0.04 &  89.5\\
            $[44]$ & 0.05 &  89.3\\
            $[63]$ & 0.05 &  89.1\\\bottomrule
        \end{tabular}        
    \caption{The outcomes of our proposed model when using different sets of $K$ in the GNN blocks.}
    \label{tab:blocks2}
\end{table}

\section{Conclusion}
\label{sec:conclusion}

In conclusion, our Multi-level Graph Convolution Neural (MLGCN) model presents a novel and efficient approach to 3D shape analysis, which is particularly for 3D object classification and 3D part segmentation from point cloud data. Our main goal was to develop a model that is lightweight and suitable for industrial and mobile applications, as most state-of-the-art models for 3D object classification can be heavy in terms of their compute and memory requirements for practical use. Our model outperforms other state-of-the-art models in terms of model size, number of operations, and number of parameters, while still achieving competitive accuracy.

Our approach uses lightweight KNN graphs shared across shallow GNN blocks to extract features from 3D point clouds at various locality levels. Our experiments demonstrate that our model can capture the relevant information in point clouds while still achieving high accuracy. 

Overall, our work represents a significant contribution to the development of efficient and effective 3D shape analysis models, with important implications for the fields of robotics, augmented reality, vision, graphics, and other industrial applications. We anticipate that our findings will motivate further research in this area, and we hope that our approach will inspire the development of even more efficient and lightweight 3D object shape abnalysis models in the future, for classification, segmentation and other vision tasks.

\bibliographystyle{unsrt}  
\bibliography{bib}

\end{document}